# EDML: A Method for Learning Parameters in Bayesian Networks

**Arthur Choi, Khaled S. Refaat and Adnan Darwiche**
Computer Science Department
University of California, Los Angeles
{*aychoi, krefaat, darwiche*}*@cs.ucla.edu*

## Abstract

We propose a method called EDML for learning MAP parameters in binary Bayesian networks under incomplete data. The method assumes Beta priors and can be used to learn maximum likelihood parameters when the priors are uninformative. EDML exhibits interesting behaviors, especially when compared to EM. We introduce EDML, explain its origin, and study some of its properties both analytically and empirically.

## 1 INTRODUCTION

We consider in this paper the problem of learning Bayesian network parameters given incomplete data, while assuming that all network variables are binary. We propose a specific method, EDML,[1] which has a similar structure and complexity to the EM algorithm (Dempster, Laird, & Rubin, 1977; Lauritzen, 1995). EDML assumes Beta priors on network parameters, allowing one to compute MAP parameters. When using uninformative priors, EDML reduces to computing maximum likelihood (ML) parameters.

EDML originated from applying an approximate inference algorithm (Choi & Darwiche, 2006) to a meta network in which parameters are explicated as variables, and on which data is asserted as evidence. The update equations of EDML resemble the ones for EM, yet EDML appears to have different convergence properties which stem from its being an inference method as opposed to a local search method. For example, we will identify a class of incomplete datasets on which EDML is guaranteed to converge immediately to an optimal solution, by simply reasoning about the behavior of its underlying inference method.

Even though EDML originates in a rather involved approximate inference scheme, its update equations can be intuitively justified independently. We therefore present EDML initially in Section 3 before delving into the details of how it was originally derived in Section 5.

Intuitively, EDML can be thought of as relying on two key concepts. The first concept is that of estimating the parameters of a single random variable given *soft observations,* i.e., observations that provide soft evidence on the values of a random variable. The second key concept behind EDML is that of interpreting the examples of an incomplete data set as providing soft observations on the random variables of a Bayesian network. As to the first concept, we also show that MAP and ML parameter estimates are unique in this case, therefore, generalizing the fundamental result which says that these estimates are unique for hard observations. This result is interesting and fundamental enough that we treat it separately in Section 4 before we move on and discuss the origin of EDML in Section 5.

We discuss some theoretical properties of EDML in Section 6, where we identify situations in which it is guaranteed to converge immediately to optimal estimates. We present some preliminary empirical results in Section 7 that corroborate some of the convergence behaviors predicted. In Section 8, we close with some concluding remarks on related and future work. We note that while we focus on binary variables here, our approach generalizes to multivalued variables as well. We will comment later on this and the reason we restricted our focus here.

## 2 TECHNICAL PRELIMINARIES

We use upper case letters ($X$) to denote variables and lower case letters ($x$) to denote their values. Variable sets are denoted by bold-face upper case letters ($\mathbf{X}$) and their instantiations by bold-face lower case letters ($\mathbf{x}$). Since our focus is on binary variables, we use $x$ (positive) and $\bar{x}$ (negative) to denote the two values of binary variable $X$. Generally, we will use $X$ to denote a variable in a Bayesian network and $\mathbf{U}$ to denote its parents. A network parameter will therefore have the general form $\theta_{x|\mathbf{u}}$, representing the probability $Pr(X=x|\mathbf{U}=\mathbf{u})$.

[1] EDML stands for **E**dge-**D**eletion **M**AP-**L**earning or **E**dge-**D**eletion **M**aximum-**L**ikelihood as it is based on an edge-deletion approximate inference algorithm that can compute MAP or maximum likelihood parameters.

Note that variable $X$ can be thought of as inducing a number of conditional random variables, denoted $X_{\mathbf{u}}$, where the values of variable $X_{\mathbf{u}}$ are drawn based on the conditional distribution $Pr(X|\mathbf{u})$. In fact, parameter estimation in Bayesian networks can be thought of as a process of estimating the distributions of these conditional random variables. Since we assume binary variables, each of these distributions can be characterized by the single parameter $\theta_{x|\mathbf{u}}$, since $\theta_{\bar{x}|\mathbf{u}} = 1-\theta_{x|\mathbf{u}}$.

We will use $\theta$ to denote the set of all network parameters. Given a network structure $G$ in which all variables are binary, our goal is to learn its parameters from an incomplete dataset, such as:

| example | $X$ | $Y$ | $Z$ |
|---|---|---|---|
| 1 | $x$ | $\bar{y}$ | ? |
| 2 | ? | $\bar{y}$ | ? |
| 3 | $\bar{x}$ | ? | $z$ |

We use $\mathcal{D}$ to denote a dataset, and $\mathbf{d}_i$ to denote an example. The dataset above has three examples, with $\mathbf{d}_3$ being the instantiation $X=\bar{x}, Z=z$.

A commonly used measure for the quality of parameter estimates $\theta$ is their likelihood, defined as:

$$L(\theta|\mathcal{D}) = \prod_{i=1}^{N} Pr_\theta(\mathbf{d}_i),$$

where $Pr_\theta$ is the distribution induced by network structure $G$ and parameters $\theta$. In the case of complete data (each example fixes the value of each variable), the ML parameters are unique. Learning ML parameters is harder when the data is incomplete, where EM is typically employed. EM starts with some initial parameters $\theta^0$, called a *seed,* and successively improves on them via iteration. EM uses the update equation:

$$\theta_{x|\mathbf{u}}^{k+1} = \frac{\sum_{i=1}^{N} Pr_{\theta^k}(x\mathbf{u}|\mathbf{d}_i)}{\sum_{i=1}^{N} Pr_{\theta^k}(\mathbf{u}|\mathbf{d}_i)},$$

which requires inference on a Bayesian network parameterized by $\theta^k$, in order to compute $Pr_{\theta^k}(x\mathbf{u}|\mathbf{d}_i)$ and $Pr_{\theta^k}(\mathbf{u}|\mathbf{d}_i)$. In fact, one run of the jointree algorithm on each distinct example is sufficient to implement an iteration of EM, which is guaranteed to never decrease the likelihood of its estimates across iterations. EM also converges to every local maxima, given that it starts with an appropriate seed. It is common to run EM with multiple seeds, keeping the best local maxima it finds. See (Darwiche, 2009; Koller & Friedman, 2009) for recent treatments on parameter learning in Bayesian networks via EM and related methods.

EM can also be used to find MAP parameters, assuming one has some priors on network parameters. The Beta distribution is commonly used as a prior on the probability of a binary random variable. In particular, the Beta for random variable $X_{\mathbf{u}}$ is specified by two exponents, $\alpha_{X_{\mathbf{u}}}$ and $\beta_{X_{\mathbf{u}}}$, leading to a density $\propto [\theta_{x|\mathbf{u}}]^{\alpha_{X_{\mathbf{u}}}-1}[1-\theta_{x|\mathbf{u}}]^{\beta_{X_{\mathbf{u}}}-1}$. It is common to assume that exponents are $> 1$ (the density is then unimodal). For MAP parameters, EM uses the update equation (see, e.g., (Darwiche, 2009)):

$$\theta_{x|\mathbf{u}}^{k+1} = \frac{\alpha_{X_{\mathbf{u}}} - 1 + \sum_{i=1}^{N} Pr_{\theta^k}(x\mathbf{u}|\mathbf{d}_i)}{\alpha_{X_{\mathbf{u}}} + \beta_{X_{\mathbf{u}}} - 2 + \sum_{i=1}^{N} Pr_{\theta^k}(\mathbf{u}|\mathbf{d}_i)}.$$

When $\alpha_{X_{\mathbf{u}}} = \beta_{X_{\mathbf{u}}} = 1$ (uninformative prior), the equation reduces to the one for computing ML parameters. When computing ML parameters, using $\alpha_{X_{\mathbf{u}}} = \beta_{X_{\mathbf{u}}} = 2$ leads to what is usually known as Laplace smoothing. This is a common technique to deal with the problem of insufficient counts (i.e., instantiations that never appear in the dataset, leading to zero probabilities and division by zero). We will indeed use Laplace smoothing in our experiments.

Our method for learning MAP and ML parameters makes heavy use of two notions: (1) the *odds* of an event, which is the probability of the event over the probability of its negation, and (2) the *Bayes factor* (Good, 1950), which is the relative change in the odds of one event, say, $X=x$, due to observing some other event, say, $\eta$. In this case, we have the odds $O(x)$ and $O(x|\eta)$, where the Bayes factor is $\kappa = O(x|\eta)/O(x)$, which is viewed as quantifying the strength of *soft evidence* $\eta$ on $X=x$. It is known that $\kappa = Pr(\eta|x)/Pr(\eta|\bar{x})$ and $\kappa \in [0,\infty]$. When $\kappa = 0$, the soft evidence reduces to hard evidence asserting $X=\bar{x}$. When $\kappa = \infty$, the soft evidence reduces to hard evidence asserting $X=x$. When $\kappa = 1$, the soft evidence is neutral and bears no information on $X=x$. A detailed discussion on the use of Bayes factors for soft evidence is given in (Chan & Darwiche, 2005).

# 3 AN OVERVIEW OF EDML

Consider Algorithm 1, which provides pseudocode for EM. EM typically starts with some initial parameters estimates, called a seed, and then iterates to monotonically improve on these estimates. Each iteration consists of two steps. The first step, Line 3, computes marginals over the families of a Bayesian network that is parameterized by the current estimates. The second step, Line 4, uses the computed probabilities to

**Algorithm 1** EM

**input:**
$G$: A Bayesian network structure
$\mathcal{D}$: An incomplete dataset $\mathbf{d}_1, \ldots, \mathbf{d}_N$
$\theta$: An initial parameterization of structure $G$
$\alpha_{X_\mathbf{u}}, \beta_{X_\mathbf{u}}$: Beta prior for each random variable $X_\mathbf{u}$

1: **while** not converged **do**
2: $Pr \leftarrow$ distribution induced by $\theta$ and $G$
3: **Compute** probabilities:

$$Pr(x\mathbf{u}|\mathbf{d}_i) \quad \text{and} \quad Pr(\mathbf{u}|\mathbf{d}_i)$$

for each family instantiation $x\mathbf{u}$ and example $\mathbf{d}_i$
4: **Update** parameters:

$$\theta_{x|\mathbf{u}} \leftarrow \frac{\alpha_{X_\mathbf{u}} - 1 + \sum_{i=1}^N Pr(x\mathbf{u}|\mathbf{d}_i)}{\alpha_{X_\mathbf{u}} + \beta_{X_\mathbf{u}} - 2 + \sum_{i=1}^N Pr(\mathbf{u}|\mathbf{d}_i)}$$

5: **return** parameterization $\theta$

**Algorithm 2** EDML

**input:**
$G$: A Bayesian network structure
$\mathcal{D}$: An incomplete dataset $\mathbf{d}_1, \ldots, \mathbf{d}_N$
$\theta$: An initial parameterization of structure $G$
$\alpha_{X_\mathbf{u}}, \beta_{X_\mathbf{u}}$: Beta prior for each random variable $X_\mathbf{u}$

1: **while** not converged **do**
2: $Pr \leftarrow$ distribution induced by $\theta$ and $G$
3: **Compute** Bayes factors:

$$\kappa^i_{x|\mathbf{u}} \leftarrow \frac{Pr(x\mathbf{u}|\mathbf{d}_i)/Pr(x|\mathbf{u}) - Pr(\mathbf{u}|\mathbf{d}_i) + 1}{Pr(\bar{x}\mathbf{u}|\mathbf{d}_i)/Pr(\bar{x}|\mathbf{u}) - Pr(\mathbf{u}|\mathbf{d}_i) + 1} \qquad (1)$$

for each family instantiation $x\mathbf{u}$ and example $\mathbf{d}_i$
4: **Update** parameters:

$$\theta_{x|\mathbf{u}} \leftarrow \underset{p}{\operatorname{argmax}}\ [p]^{\alpha_{X_\mathbf{u}}-1}[1-p]^{\beta_{X_\mathbf{u}}-1}\prod_{i=1}^N [\kappa^i_{x|\mathbf{u}} \cdot p - p + 1] \qquad (2)$$

5: **return** parameterization $\theta$

update the network parameters. The process continues until some convergence criterion is met. The main point here is that the *computation* on Line 3 can be implemented by a single run of the jointree algorithm, while the *update* on Line 4 is immediate.

Consider now Algorithm 2, which provides pseudocode for EDML, to be contrasted with the one for EM. The two algorithms clearly have the same overall structure. That is, EDML also starts with some initial parameters estimates, called a seed, and then iterates to update these estimates. Each iteration consists of two steps. The first step, Line 3, computes Bayes factors using a Bayesian network that is parameterized by the current estimates. The second step, Line 4, uses the computed Bayes factors to update network parameters. The process continues until some convergence criterion is met. Much like EM, the *computation* on Line 3 can be implemented by a single run of the jointree algorithm. Unlike EM, however, the *update* on Line 4 is not immediate as it involves solving an optimization problem, albeit a simple one. Aside from this optimization task, EM and EDML have the same computational complexity.

We next explain the two concepts underlying EDML and how they lead to the equations of Algorithm 2.

## 3.1 ESTIMATION FROM SOFT OBSERVATIONS

Consider a random variable $X$ with values $x$ and $\bar{x}$, and suppose that we have $N > 0$ independent observations of $X$, with $N_x$ as the number of positive observations. It is well known that the ML parameter estimates for random variable $X$ are unique in this case and characterized by $\theta_x = N_x/N$. If one further assumes a Beta prior with exponents $\alpha$ and $\beta$ that are $\geq 1$, it is also known that the MAP parameter estimates are unique and characterized by $\theta_x = \frac{N_x+\alpha-1}{N+\alpha+\beta-2}$.

Consider now a more general problem in which the observations are soft in that they only provide soft evidence on the values of random variable $X$. That is, each soft observation $\eta_i$ is associated with a Bayes factor $\kappa^i_x = O(x|\eta_i)/O(x)$ which quantifies the evidence that $\eta_i$ provides on having observed the value $x$ of variable $X$. We will show later that the ML estimates remain unique in this more general case, if at least one of the soft observations is not trivial (i.e., with Bayes factor $\kappa^i_x \neq 1$). Moreover, we will show that the MAP estimates are also unique assuming a Beta prior with exponents $\geq 1$. In particular, we will show that the unique MAP estimates are characterized by Equation 2 of Algorithm 2. Further, we will show that the unique ML estimates are characterized by the same equation while using a Beta prior with exponents $= 1$. This is the first key concept that underlies our proposed algorithm for estimating ML and MAP parameters in a binary Bayesian network.

## 3.2 EXAMPLES AS SOFT OBSERVATIONS

The second key concept underlying EDML is to interpret each example $\mathbf{d}_i$ in a dataset as providing a soft observation on each random variable $X_\mathbf{u}$. As mentioned earlier, soft observations are specified by Bayes factors and, hence, one needs to specify the Bayes factor $\kappa^i_{x|\mathbf{u}}$ that example $\mathbf{d}_i$ induces on random variable $X_{\mathbf{u}}$. EDML uses Equation 1 for this purpose, which will be derived in Section 5. We next consider a few special cases of this equation to highlight its behavior.

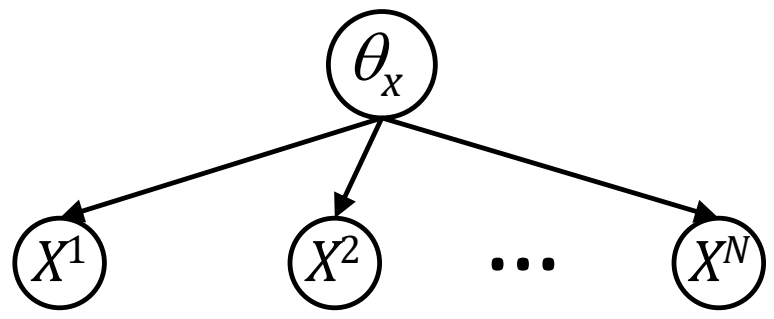

Figure 1: Estimation given independent observations.

Consider first the case in which example $\mathbf{d}_i$ implies parent instantiation $\mathbf{u}$ (i.e., the parents $\mathbf{U}$ of variable $X$ are instantiated to $\mathbf{u}$ in example $\mathbf{d}_i$). In this case, Equation 1 reduces to $\kappa^i_{x|\mathbf{u}} = \frac{O(x|\mathbf{u},\mathbf{d}_i)}{O(x|\mathbf{u})}$, which is the relative change in the odds of $x$ given $\mathbf{u}$ due to conditioning on example $\mathbf{d}_i$. Note that for root variables $X$, which have no parents $\mathbf{U}$, Equation 1 further reduces to $\kappa^i_x = \frac{O(x|\mathbf{d}_i)}{O(x)}$.

The second case we consider is when example $\mathbf{d}_i$ is inconsistent with parent instantiation $\mathbf{u}$. In this case, Equation 1 reduces to $\kappa^i_{x|\mathbf{u}} = 1$, which amounts to neutral evidence. Hence, example $\mathbf{d}_i$ is irrelevant to estimating the distribution of variable $X_{\mathbf{u}}$ in this case, and will be ignored by EDML.

The last special case of Equation 1 we shall consider is when the example $\mathbf{d}_i$ is complete; that is, it fixes the value of each variable. In this case, one can verify that $\kappa^i_{x|\mathbf{u}} \in \{0, 1, \infty\}$ and, hence, the example can be viewed as providing either neutral or hard evidence on each random variable $X_{\mathbf{u}}$. Thus, an example will provide soft observations on variables only when it is incomplete (i.e., missing some values). Otherwise, it is either irrelevant to, or provides a hard observation on, each variable $X_{\mathbf{u}}$.

In the next section, we prove Equation 2 of Algorithm 2. In Section 5, we discuss the origin of EDML, where we go on and derive Equation 1 of Algorithm 2.

# 4 ESTIMATION FROM SOFT OBSERVATIONS

Consider a binary variable $X$. Figure 1 depicts a network where $\theta_x$ is a parameter representing $Pr(X = x)$ and $X^1, \ldots, X^N$ are independent observations of $X$. Suppose further that we have a Beta prior on parameter $\theta_x$ with exponents $\alpha \geq 1$ and $\beta \geq 1$. A standard estimation problem is to assume that we know the values of these observations and then estimate the parameter $\theta_x$. We now consider a variant on this problem, in which we only have soft evidence $\eta_i$ about each observation, whose strength is quantified by a Bayes factor $\kappa^i_x = O(x|\eta_i)/O(x)$. Here, $\kappa^i_x$ represents the change in odds that the $i$-th observation is positive due to evidence $\eta_i$. We will refer to $\eta_i$ as a *soft observation* on variable $X$, and our goal in this section is to compute (and optimize) the posterior density on parameter $\theta_x$ given these soft observations $\eta_1, \ldots, \eta_N$.

We first consider the likelihood:

$$
\begin{aligned}
&Pr(\eta_1, \ldots, \eta_N|\theta_x) = \textstyle\prod_{i=1}^N Pr(\eta_i|\theta_x) \\
&= \textstyle\prod_{i=1}^N [Pr(\eta_i|x, \theta_x)Pr(x|\theta_x) + Pr(\eta_i|\bar{x}, \theta_x)Pr(\bar{x}|\theta_x)] \\
&= \textstyle\prod_{i=1}^N [Pr(\eta_i|x)\theta_x + Pr(\eta_i|\bar{x})(1 - \theta_x)] \\
&\propto \textstyle\prod_{i=1}^N [\kappa^i_x \cdot \theta_x - \theta_x + 1].
\end{aligned}
$$

The last step follows because $\kappa^i_x = O(x|\eta_i)/O(x) = Pr(\eta_i|x)/Pr(\eta_i|\bar{x})$. The posterior density is then:

$$
\begin{aligned}
&\rho(\theta_x|\eta_1, \ldots, \eta_N) \propto \rho(\theta_x)Pr(\eta_1, \ldots, \eta_N|\theta_x) \\
&\propto [\theta_x]^{\alpha-1}[1 - \theta_x]^{\beta-1} \textstyle\prod_{i=1}^N [\kappa^i_x \cdot \theta_x - \theta_x + 1].
\end{aligned}
$$

This is exactly Equation 2 of Algorithm 2 assuming we replace the random variable $X$ with the conditional random variable $X_{\mathbf{u}}$.[2]

The second derivative of the log posterior is

$$
-\frac{\alpha - 1}{[\theta_x]^2} - \frac{\beta - 1}{[1 - \theta_x]^2} - \sum_i \left[\frac{(\kappa^i_x - 1)}{(\kappa^i_x - 1)\theta_x + 1}\right]^2
$$

which is strictly negative when $\kappa^i_x \neq 1$ for at least one $i$. This remains true when $\alpha = \beta = 1$. Hence, both the likelihood function and the posterior density are strictly log-concave and therefore have unique modes. This means that both ML and MAP parameter estimates are unique in the case of soft, independent observations, which generalizes the uniqueness result for hard, independent observations on a variable $X$.

# 5 THE ORIGIN OF EDML

This section reveals the technical origin of EDML, showing how Equation 1 of Algorithm 2 is derived, and providing the basis for the overall structure of EDML as spelled out in Algorithm 2.

EDML originated from an approximation algorithm for computing MAP parameters in a *meta network*. Figure 2 depicts an example meta network in which parameters are represented explicitly as nodes (Darwiche, 2009). In particular, for each conditional random variable $X_{\mathbf{u}}$ in the original Bayesian network, called the *base network*, we have a node $\theta_{x|\mathbf{u}}$ in the meta network which represents a parameter that characterizes the distribution of this random variable. Moreover, the meta network includes enough instances of the base network to allow the assertion of each example $\mathbf{d}_i$ as evidence on one of these instances. Assuming that $\theta$ is an instantiation of all parameter variables, and $\mathcal{D}$ is a dataset, MAP estimates are then:

$$\theta^\star = \underset{\theta}{\operatorname{argmax}}\, \rho(\theta|\mathcal{D}),$$

where $\rho$ is the density induced by the meta network.

[2] The case of $\kappa^i_x = \infty$ needs to be handled carefully in Equation 2. First note that $\kappa^i_x = \infty$ iff $Pr(\eta_i|\bar{x}) = 0$ in the derivation of this equation. In this case, the term $Pr(\eta_i|x)\theta_x + Pr(\eta_i|\bar{x})(1-\theta_x)$ equals $c \cdot \theta_x$ for some constant $c \in (0, 1]$. Since the value of Equation 2 does not depend on constant $c$, we will assume $c = 1$. Hence, when $\kappa^i_x = \infty$, the term $[\kappa^i_x \cdot \theta_x - \theta_x + 1]$ evaluates to $\theta_x$ by convention.

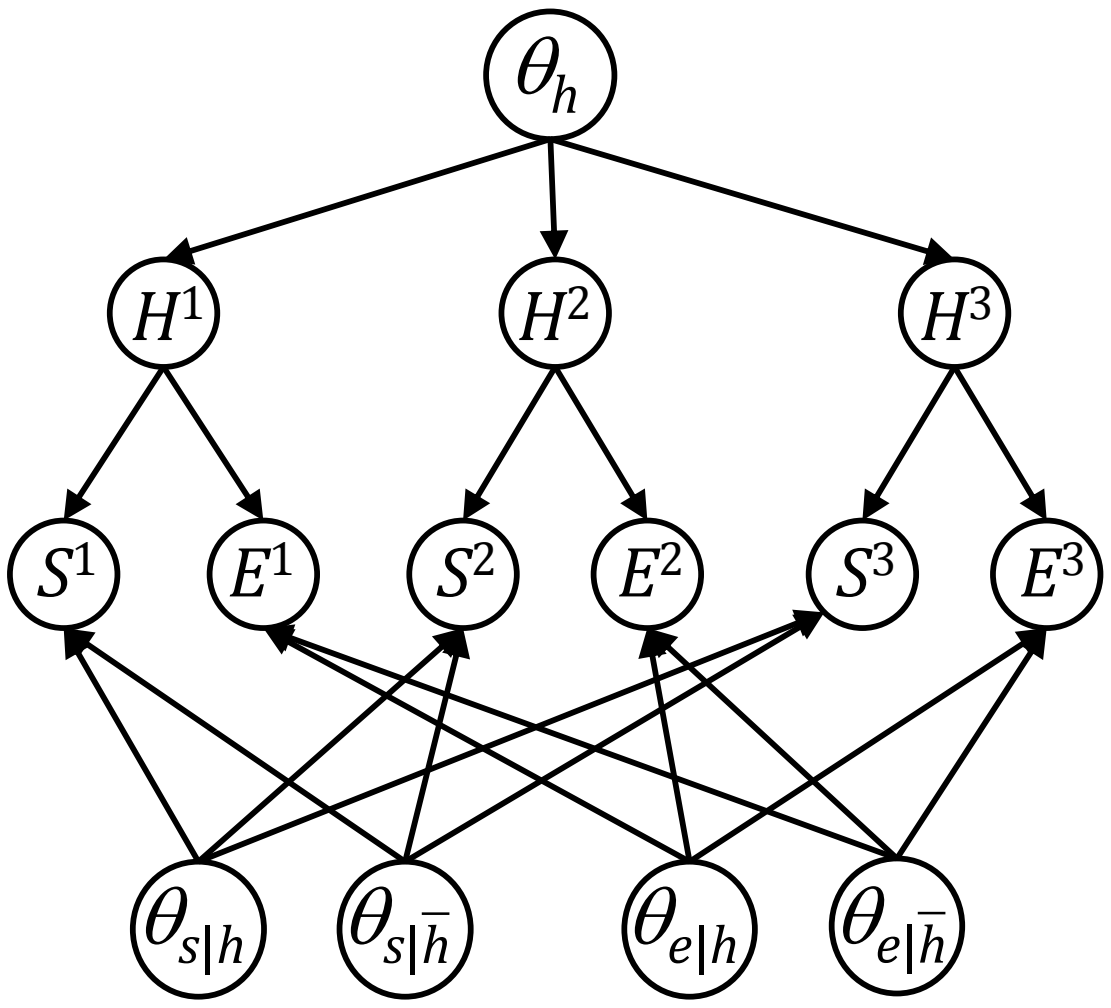

Figure 2: A meta network induced from a base network $S \longleftarrow H \longrightarrow E$. The CPTs here are based on standard semantics; see, e.g., (Darwiche, 2009, Ch. 18).

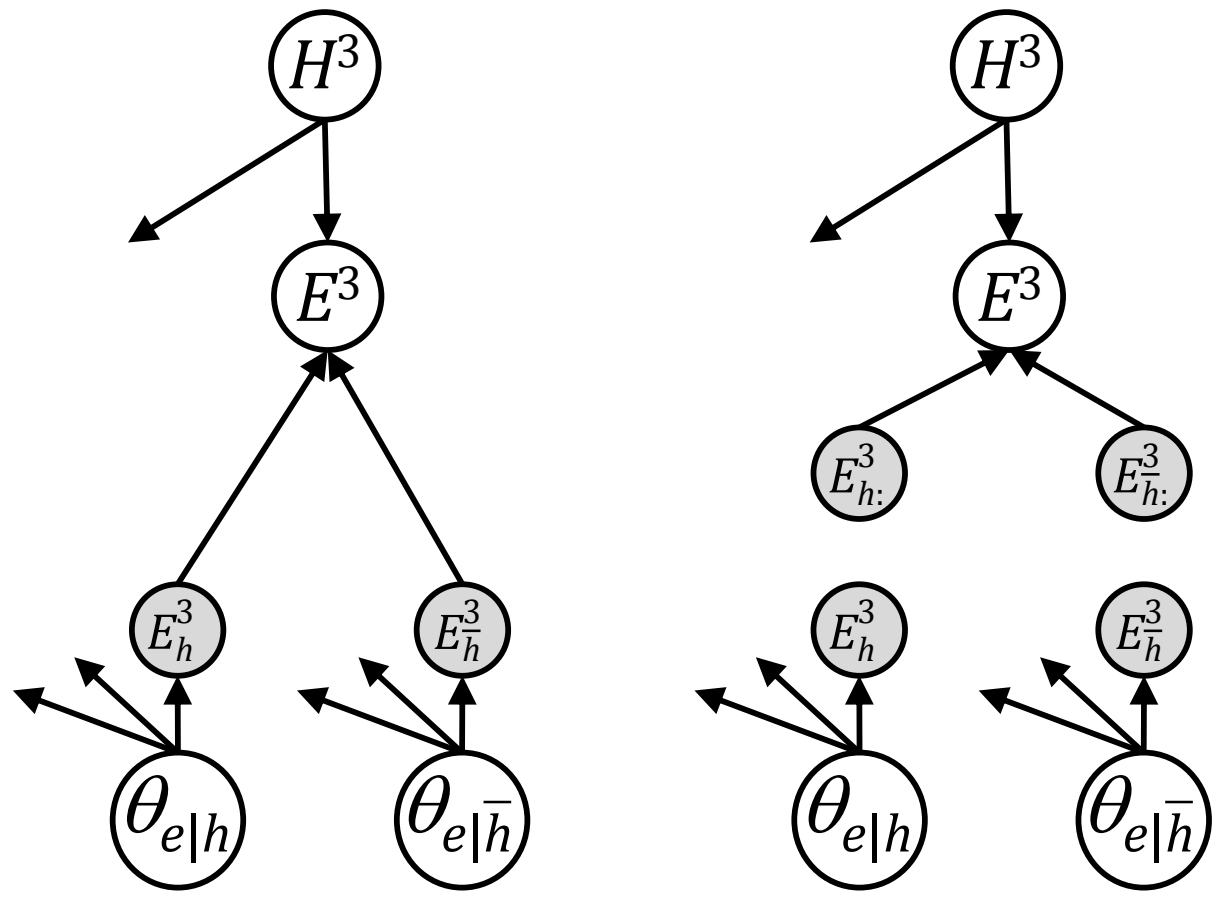

(a) Adding generators (b) Deleting copy edges

Figure 3: Introducing generators into a meta network and then deleting copy edges from the resulting meta network, which leads to introducing clones.

Computing MAP estimates exactly is usually prohibitive due to the structure of the meta network. We therefore use the technique of edge deletion (Choi & Darwiche, 2006), which formulates approximate inference as exact inference on a simplified network that is obtained by deleting edges from the original network. The technique compensates for these deletions by introducing auxiliary parameters whose values must be chosen carefully (and usually iteratively) in order to improve the quality of approximations obtained from the simplified network. EDML is the result of making a few specific choices for deleting edges and for choosing values for the auxiliary parameters introduced, which we explain next.

### 5.1 INTRODUCING GENERATORS

Let $X^i$ denote the instance of variable $X$ in the base network corresponding to example $\mathbf{d}_i$. The first choice of EDML is that for each edge $\theta_{x|\mathbf{u}} \longrightarrow X^i$ in the meta network, we introduce a *generator variable* $X^i_{\mathbf{u}}$, leading to the pair of edges $\theta_{x|\mathbf{u}} \longrightarrow X^i_{\mathbf{u}} \longrightarrow X^i$. Figure 3(a) depicts a fragment of the meta network in Figure 2, in which we introduced two generator variables for edges $\theta_{e|h} \longrightarrow E^3$ and $\theta_{e|\bar{h}} \longrightarrow E^3$, leading to $\theta_{e|h} \longrightarrow E^3_h \longrightarrow E^3$ and $\theta_{e|\bar{h}} \longrightarrow E^3_{\bar{h}} \longrightarrow E^3$.

Variable $X^i_{\mathbf{u}}$ is meant to generate values of variable $X^i$ according to the distribution specified by parameter $\theta_{x|\mathbf{u}}$. Hence, the conditional distribution of a generator $X^i_{\mathbf{u}}$ is such that $Pr(x^i_{\mathbf{u}}|\theta_{x|\mathbf{u}}) = \theta_{x|\mathbf{u}}$. Moreover, the CPT of variable $X^i$ is set to ensure that variable $X^i$ copies the value of generator $X^i_{\mathbf{u}}$ if and only if the parents of $X^i$ take on the value $\mathbf{u}$. That is, the CPT of variable $X^i$ acts as a selector that chooses a particular generator $X^i_{\mathbf{u}}$ to copy from, depending on the values of its parents $\mathbf{U}$. For example, in Figure 3(a), when parent $H^3$ takes on its positive value $h$, variable $E^3$ copies the value of generator $E^3_h$. When parent $H^3$ takes on its negative value $\bar{h}$, variable $E^3$ copies the value of generator $E^3_{\bar{h}}$.

Adding generator variables does not change the meta network as it continues to have the same density over the original variables. Yet, generators are essential to the derivation of EDML as they will be used for interpreting data examples as soft observations.

### 5.2 DELETING COPY EDGES

The second choice made by EDML is that we only delete edges of the form $X^i_{\mathbf{u}} \longrightarrow X^i$ from the augmented meta network, which we shall call *copy edges*. Figure 3(b) depicts an example in which we have deleted

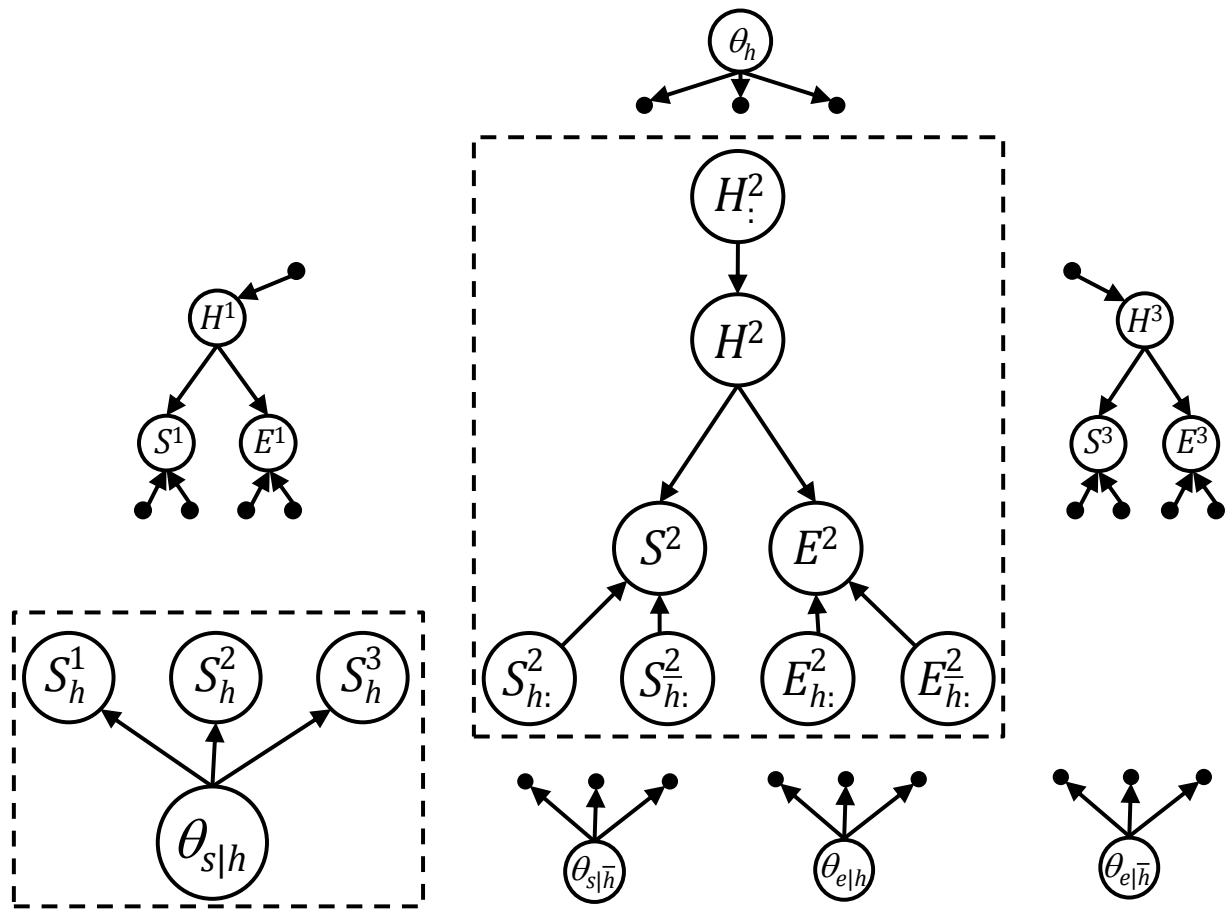

Figure 4: An edge-deleted network obtained from the meta network in Figure 2 found by: (1) adding generator variables, (2) deleting copy edges, and (3) adding cloned generators. The figure highlights the island for example $\mathbf{d}_2$, and the island for parameter $\theta_{s|h}$.

the two copy edges from Figure 3(a).

Note here the addition of another auxiliary variable $X^i_{\mathbf{u}:}$, called a *clone,* for each generator $X^i_{\mathbf{u}}$. The addition of clones is mandated by the edge deletion framework. Moreover, if the CPT of clone $X^i_{\mathbf{u}:}$ is chosen carefully, it can compensate for the parent-to-child information lost when deleting edge $X^i_{\mathbf{u}} \longrightarrow X^i$. We will later see how EDML sets these CPTs. The other aspect of compensating for a deleted edge is to specify soft evidence on each generator $X^i_{\mathbf{u}}$. This is also mandated by the edge deletion framework, and is meant to compensate for the child-to-parent information lost when deleting edge $X^i_{\mathbf{u}} \longrightarrow X^i$. We will later see how EDML sets this soft evidence as well, which effectively completes the specification of the algorithm. We prelude this specification, however, by making some further observations about the structure of the meta network after edge deletion.

### 5.3 PARAMETER & EXAMPLE ISLANDS

Consider the network in Figure 4, which is obtained from the meta network in Figure 2 according to the edge-deletion process indicated earlier.

The edge-deleted network contains a set of disconnected structures, called *islands.* Each island belongs to one of two classes: a *parameter island* for each network parameter $\theta_{x|\mathbf{u}}$ and an *example island* for each example $\mathbf{d}_i$ in the dataset. Figure 4 provides the full details for one example island and one parameter island. Note that each parameter island corresponds to a Naive Bayes structure, with parameter $\theta_{x|\mathbf{u}}$ as the root and generators $X^i_{\mathbf{u}}$ as children. When soft evidence is asserted on these generators, we get the estimation problem we treated in Section 4.

EDML can now be fully described by specifying (1) the soft evidence on each generator $X^i_{\mathbf{u}}$ in a parameter island, and (2) the CPT of each clone $X^i_{\mathbf{u}:}$ in an example island. These specifications are given next.

### 5.4 CHILD-TO-PARENT COMPENSATION

The edge deletion approach suggests the following soft evidence on generators $X^i_{\mathbf{u}}$, specified as Bayes factors:

$$\kappa^i_{x|\mathbf{u}} = \frac{O(x^i_{\mathbf{u}:}|\mathbf{d}_i)}{O(x^i_{\mathbf{u}:})} = \frac{Pr^i(\mathbf{d}_i|x^i_{\mathbf{u}:})}{Pr^i(\mathbf{d}_i|\bar{x}^i_{\mathbf{u}:})}, \tag{3}$$

where $Pr^i$ is the distribution induced by the island of example $\mathbf{d}_i$. We will now show that this equation simplifies to Equation 1 of Algorithm 2.

Suppose that we marginalize all clones $X^i_{\mathbf{u}:}$ from the island of example $\mathbf{d}_i$, leading to a network that induces a distribution $Pr$. The new network has the following properties. First, it has the same structure as the base network. Second, $Pr(x|\mathbf{u}) = Pr^i(x^i_{\mathbf{u}:})$, which means that the CPTs of clones in example islands correspond to parameters in the base network. Finally, if we use $\overline{\mathbf{u}}$ to denote the disjunction of all parent instantiations excluding $\mathbf{u}$, we get:

$$\begin{aligned}\kappa^i_{x|\mathbf{u}} &= \frac{Pr^i(\mathbf{d}_i|x^i_{\mathbf{u}:})}{Pr^i(\mathbf{d}_i|\bar{x}^i_{\mathbf{u}:})} \\ &= \frac{Pr(\mathbf{d}_i|x\mathbf{u})Pr(\mathbf{u}) + Pr(\mathbf{d}_i|\overline{\mathbf{u}})Pr(\overline{\mathbf{u}})}{Pr(\mathbf{d}_i|\bar{x}\mathbf{u})Pr(\mathbf{u}) + Pr(\mathbf{d}_i|\overline{\mathbf{u}})Pr(\overline{\mathbf{u}})} \\ &= \frac{Pr(x\mathbf{u}|\mathbf{d}_i)/Pr(x|\mathbf{u}) - Pr(\mathbf{u}|\mathbf{d}_i) + 1}{Pr(\bar{x}\mathbf{u}|\mathbf{d}_i)/Pr(\bar{x}|\mathbf{u}) - Pr(\mathbf{u}|\mathbf{d}_i) + 1}.\end{aligned}$$

This is exactly Equation 1 of Algorithm 2. Hence, we can evaluate Equation 3 by evaluating Equation 1 on the base network, as long as we seed the base network with parameters that correspond to the CPTs of clones in an example island.

### 5.5 PARENT-TO-CHILD COMPENSATION

We now complete the derivation of EDML by showing how it specifies the CPTs of clones in example islands, which are needed for computing soft evidence as in the previous section.

In a nutshell, EDML assumes an initial value of these CPTs, typically chosen randomly. Given these CPTs, example islands will be fully specified and EDML will compute soft evidence as given by Equation 3. The computed soft evidence is then injected on the generators of parameter islands, leading to a full specification of these islands. EDML will then estimate parameters by solving an exact optimization problem on each parameter island as shown in Section 4. The estimated parameters are then used as the new values of CPTs for clones in example islands. This process repeats until convergence.

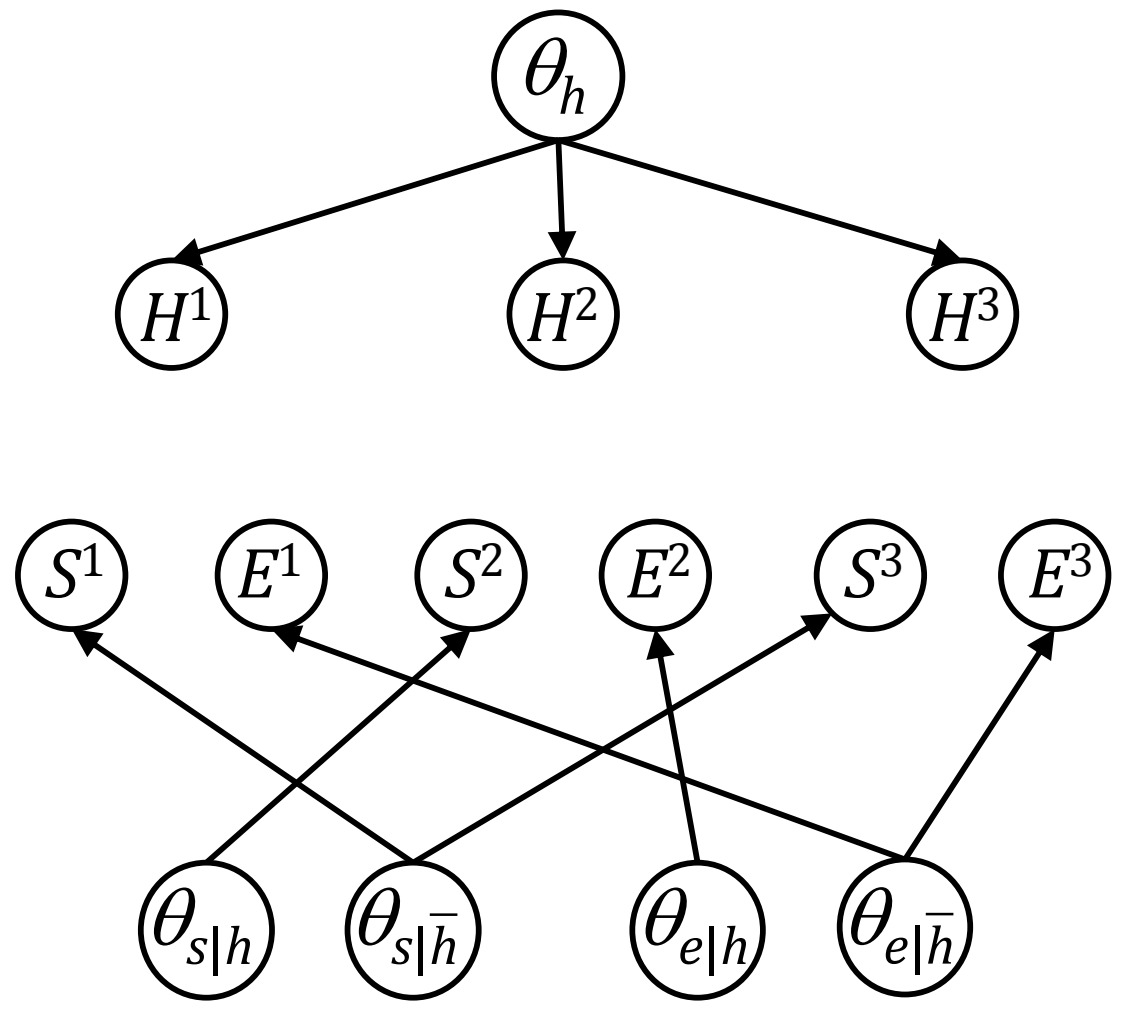

Figure 5: A pruning of the meta network in Figure 2 given $H^1 = \bar{h}$, $H^2 = h$ and $H^3 = \bar{h}$.

We have shown in the previous section that the CPTs of clones are in one-to-one correspondence with the parameters of the base network. We have also shown that soft evidence, as given by Equation 3, can be computed by evaluating Equation 1 of Algorithm 2 (with parameters $\theta$ corresponding to the CPTs of clones in an example island). EDML takes advantage of this correspondence, leading to the simplified statement spelled out in Algorithm 2.

## 6 SOME PROPERTIES OF EDML

Being an approximate inference method, one can sometimes identify good behaviors of EDML by identifying situations under which the underlying inference algorithm will produce high quality approximations. We provide a result in this section that illustrates this point in the extreme, where EDML is guaranteed to return optimal estimates and in only one iteration. Our result relies on the following observation about parameter estimation via inference on a meta network.

When the parents $\mathbf{U}$ of a variable $X$ are observed to $\mathbf{u}$ in an example $\mathbf{d}_i$, all edges $\theta_{x|\mathbf{u}\star} \longrightarrow X^i$ in the meta network become superfluous and can be pruned, except for the one edge that satisfies $\mathbf{u}\star = \mathbf{u}$. Moreover, edges outgoing from observed nodes can also be pruned from a meta network. Suppose now that the parents of each variable are observed in a dataset. After pruning edges as indicated earlier, each parameter variable $\theta_{x|\mathbf{u}}$ will end up being the root of an isolated naive Bayes structure that has some variables $X^i$ as its children (those whose parents are instantiated to $\mathbf{u}$ in example $\mathbf{d}_i$). Figure 5 depicts the result of such pruning in the meta network of Figure 2, given a dataset with $H^1 = \bar{h}$, $H^2 = h$ and $H^3 = \bar{h}$.

The above observation implies that when the parents of each variable are observed in a dataset, parameters can be estimated independently. This leads to the following well known result.

**Proposition 1** *When the dataset is complete, the ML estimate for parameter $\theta_{x|\mathbf{u}}$ is unique and given by $\mathcal{D}\#(x\mathbf{u})/\mathcal{D}\#(\mathbf{u})$, where $\mathcal{D}\#(x\mathbf{u})$ is the number of examples containing $x\mathbf{u}$ and $\mathcal{D}\#(\mathbf{u})$ is the number of examples containing $\mathbf{u}$.*

It is well known that EM returns such estimates and in only one iteration (i.e., independently of its seed). The following more general result is also implied by our earlier observation.

**Proposition 2** *When only leaf variables have missing values in a dataset, the ML estimate for each parameter $\theta_{x|\mathbf{u}}$ is unique and given by $\mathcal{D}\#(x\mathbf{u})/\mathcal{D}^+\#(\mathbf{u})$. Here, $\mathcal{D}^+\#(\mathbf{u})$ is the number of examples containing $\mathbf{u}$ and in which $X$ is observed.*

We can now prove the following property of EDML, which is not satisfied by EM, as we show next.

**Theorem 1** *When only leaf variables have missing values in a dataset, EDML returns the unique ML estimates given by Proposition 2 and in only one iteration.*

**Proof** Consider an example $\mathbf{d}_i$ that fixes the values of parents $\mathbf{U}$ for variable $X$ and consider Equation 1. First, $\kappa^i_{x|\mathbf{u}} = 1$ iff example $\mathbf{d}_i$ is inconsistent with $\mathbf{u}$ or does not set the value of $X$. Next, $\kappa^i_{x|\mathbf{u}} = 0$ iff example $\mathbf{d}_i$ contains $\bar{x}\mathbf{u}$. Finally, $\kappa^i_{x|\mathbf{u}} = \infty$ iff example $\mathbf{d}_i$ contains $x\mathbf{u}$. Moreover, these values are independent of the EDML seed so the algorithm converges in one iteration. Given these values of the Bayes factors, Equation 2 leads to the estimate of Proposition 2. □

We have a number of observations about this result. First, since Proposition 1 is implied by Proposition 2, EDML returns the unique ML estimates in only one iteration when the dataset is complete (just like EM). Next, when only the values of leaf variables are missing in a dataset, Proposition 2 says that there is a unique ML estimate for each network parameter. Moreover,

Theorem 1 says that EDML returns these unique estimates and in only one iteration. Finally, Theorem 1 does not hold for EM. In particular, one can show that under the conditions of this theorem, an EM iteration will update its current parameter estimates $\theta$ and return the following estimates for $\theta_{x|\mathbf{u}}$:

$$\frac{\mathcal{D}\#(x\mathbf{u}) + \mathcal{D}^{-}\#(\mathbf{u})Pr(x|\mathbf{u})}{\mathcal{D}\#(\mathbf{u})}.$$

Here, $\mathcal{D}^{-}\#(\mathbf{u})$ is the number of examples that contain $\mathbf{u}$ and in which the value of $X$ is missing. This next estimate clearly depends on the current parameter estimates. As a result, the behavior of EM will depend on its initial seed, unlike EDML.

When only the values of leaf variables are missing, there is a unique optimal solution as shown by Proposition 2. Since EM is known to converge to a local optimum, it will eventually return the optimal estimates as well, but possibly after some number of iterations. In this case, the difference between EM and EDML is simply in the speed of convergence.

Theorem 1 clearly suggests better convergence behavior of EDML over EM in some situations. We next present initial experiments supporting this suggestion.

# 7 MORE ON CONVERGENCE

We highlight now a few empirical properties of EDML. In particular, we show how EDML can sometimes find higher quality estimates than EM, in fewer iterations and also in less time.

We highlight different types of relative convergence behavior in Figure 6, which depicts example runs on a selection of networks: `spect`, `win95pts`, `emdec6g`, and `tcc4e`. Network `spect` is a naive Bayes network induced from a dataset in the UCI ML repository, with 1 class variable and 22 attributes. Network `win95pts` (76 variables) is an expert system for printer troubleshooting in Windows 95. Networks `emdec6g` (168 variables) and `tcc4e` (98 variables) are noisy-or networks for diagnosis (courtesy of HRL Laboratories).

We simulated datasets of size $2^k$, using the original CPT parameters of the respective networks, and then used EDML and EM to learn new parameters for a network with the same structure. We assumed that certain variables were hidden (latent); in Figure 6, we randomly chose $\frac{1}{4}$ of the variables to be hidden. Hidden nodes are of particular interest to EM, because it has been observed that local extrema and convergence rates can be problematic for EM here; see, for example (Elidan & Friedman, 2005; Salakhutdinov, Roweis, & Ghahramani, 2003).

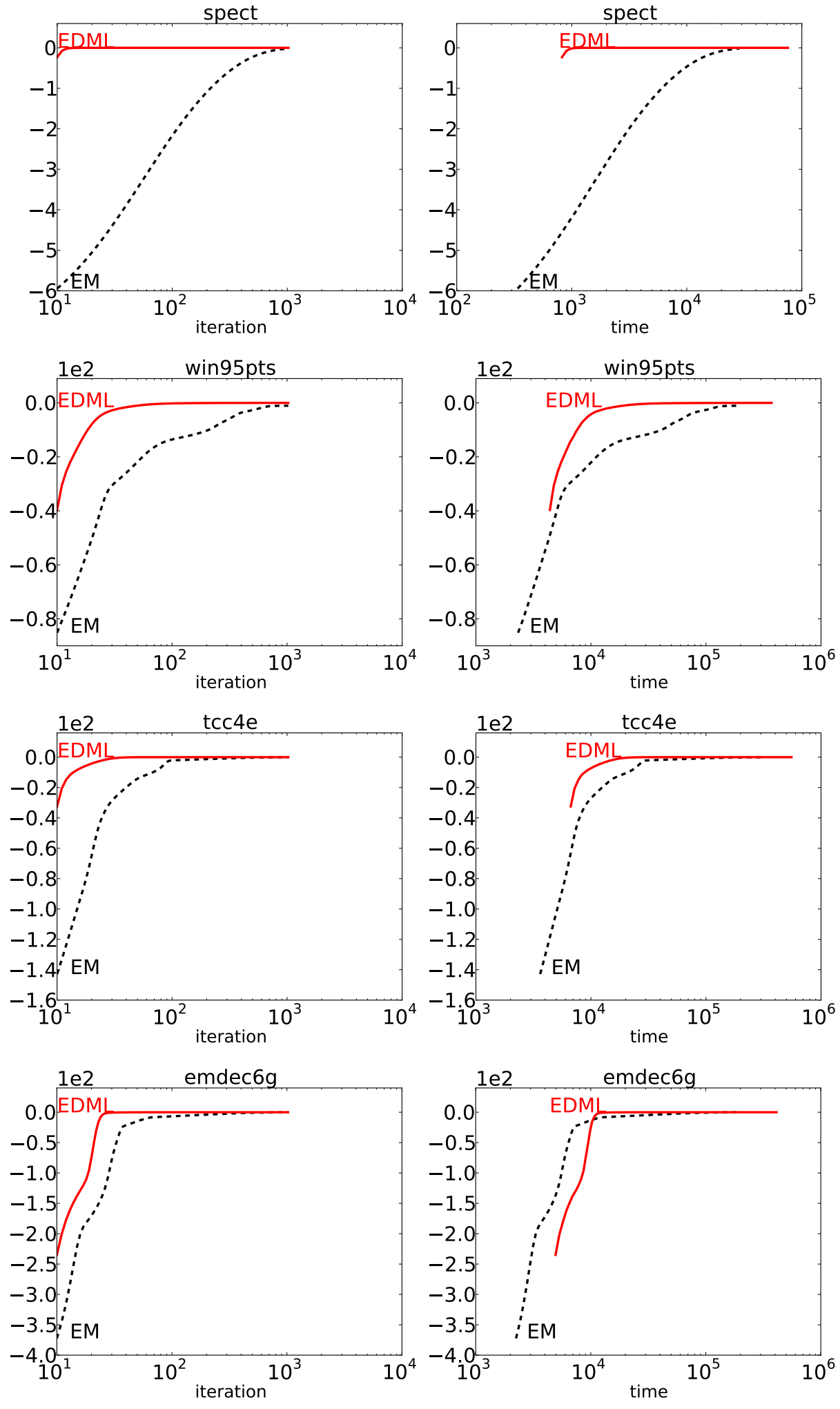

Figure 6: Quality of parameter estimates over iterations (left column) and time (right column). Going right on the $x$-axis, we have increasing iterations and time. Going up on the $y$-axis, we have increasing quality of parameter estimates. EDML is depicted with a solid red line, and EM with a dashed black line.

In Figure 6, each plot represents a simulated data set of size $2^{10}$, where EDML and EM have been initialized with the same random parameter seeds. Both algorithms were run for a fixed number of iterations, 1024 in this case, and we observed the quality of the parameter estimates found, with respect to the log posterior probability (which has been normalized so that the maximum log probability observed is 0.0). We assumed a Beta prior with exponents 2. EDML damped its parameter updates by a factor of $\frac{1}{2}$, which is typical for (loopy) belief propagation algorithms.[3]

[3]The simple bisection method suffices for the optimization sub-problem in EDML for binary Bayesian networks. In our current implementation, we used the conjugate gradient method, with a convergence threshold of $10^{-8}$.

In the left column of Figure 6, we evaluated the quality of estimates over iterations of EDML and EM. In these examples, EDML (represented by a solid red line) tended to have better quality estimates from iteration to iteration (curves that are higher are better), and further managed to find them in fewer iterations (curves to the left are faster).[4] This is most dramatic in network `spect`, where EDML appears to have converged almost immediately, whereas EM spent a significant number of iterations to reach estimates of comparable quality. As most nodes hidden in network `spect` were leaf nodes, this may be expected due to the considerations from the previous section.

In the right column of Figure 6, we evaluated the quality of estimates, now in terms of *time*. We remark again that procedurally, EDML and EM are very similar, and each algorithm needs only one evaluation of the jointree algorithm per distinct example in the data set (per iteration). EDML solves an optimization problem per distinct example, whereas EM has a closed-form update equation in the corresponding step (Line 4 in Algorithms 1 and 2). Although this optimization problem is a simple one, EDML does require more time per iteration than EM. The right column of Figure 6 suggests that EDML can still find better estimates faster, especially in the cases where EDML has converged in significantly fewer iterations. In network `emdec6g`, we find that although EDML appeared to converge in fewer iterations, EM was able to find better estimates in less time. We anticipate in larger networks with higher treewidth, the time spent in the simple optimization sub-problem will be dominated by the time to perform jointree propagation.

We also performed experiments on networks learned from binary haplotype data (Elidan & Gould, 2008), which are networks with bounded treewidth. Here, we simulated data sets of size $2^{10}$, where we again randomly selected $\frac{1}{4}$ of the variables to be hidden. We further ran EDML and EM for a fixed number of iterations (512, here). For each of the 74 networks available, we ran EDML and EM with 3 random seeds, for a total of 222 cases. In Figure 7, we highlight a selection of the runs we performed, to illustrate examples of relative convergence behaviors. Again, in the first row, we see a case where EDML identifies better estimates in fewer iterations and less time. In the next two rows, we highlight two cases where EDML appears to converge to a superior fixed point than the one that EM appears to converge to. In the last row, we highlight an instance where EM instead converges to a superior estimate. In Figure 8, we compare the estimates of EDML and EM at each iteration, computing the percentage of the $74 \times 3 = 222$ cases considered, where EDML had estimates no worse than those found by EM. In this set of experiments, the estimates identified by EDML are clearly superior (or at least, no worse in most cases), when compared to EM.

[4]We omit the results of the first 10 iterations as initial parameter estimates are relatively poor, which make the plots difficult to read.

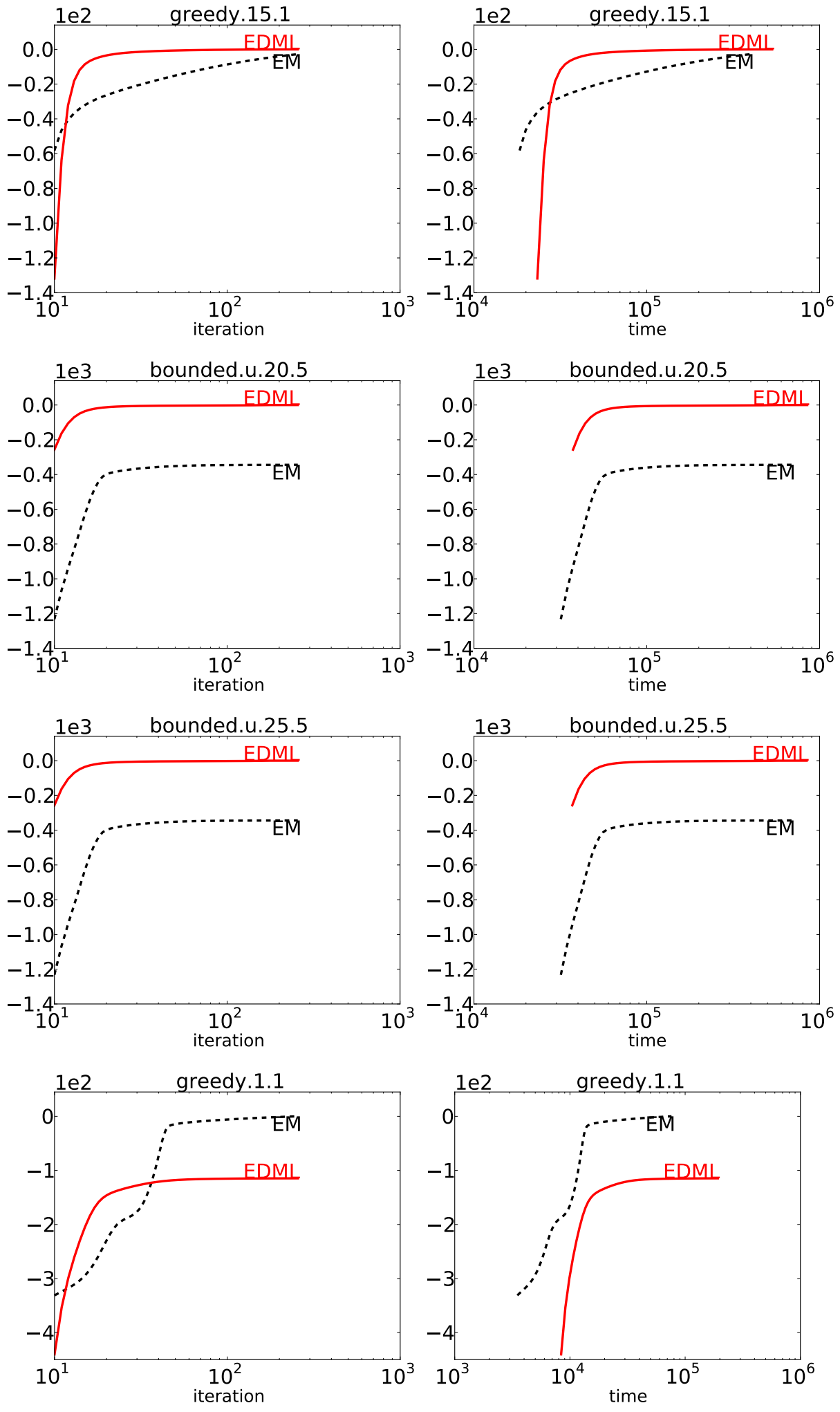

Figure 7: Quality of parameter estimates over iterations (left column) and time (right column). Going right on the $x$-axis, we have increasing iterations and time. Going up the $y$-axis, we have increasing quality of parameter estimates. EDML is depicted with a solid red line, and EM with a dashed black line.

We remark however, that when both algorithms are given enough iterations to converge, we have observed that the quality of the estimates found by both algorithms are often comparable. This is evident in Figure 6, for example. The analysis from the previous section indicates however that there are (very specialized) situations where EDML would be clearly preferred over EM. One subject of future study is the identification of situations and applications where EDML

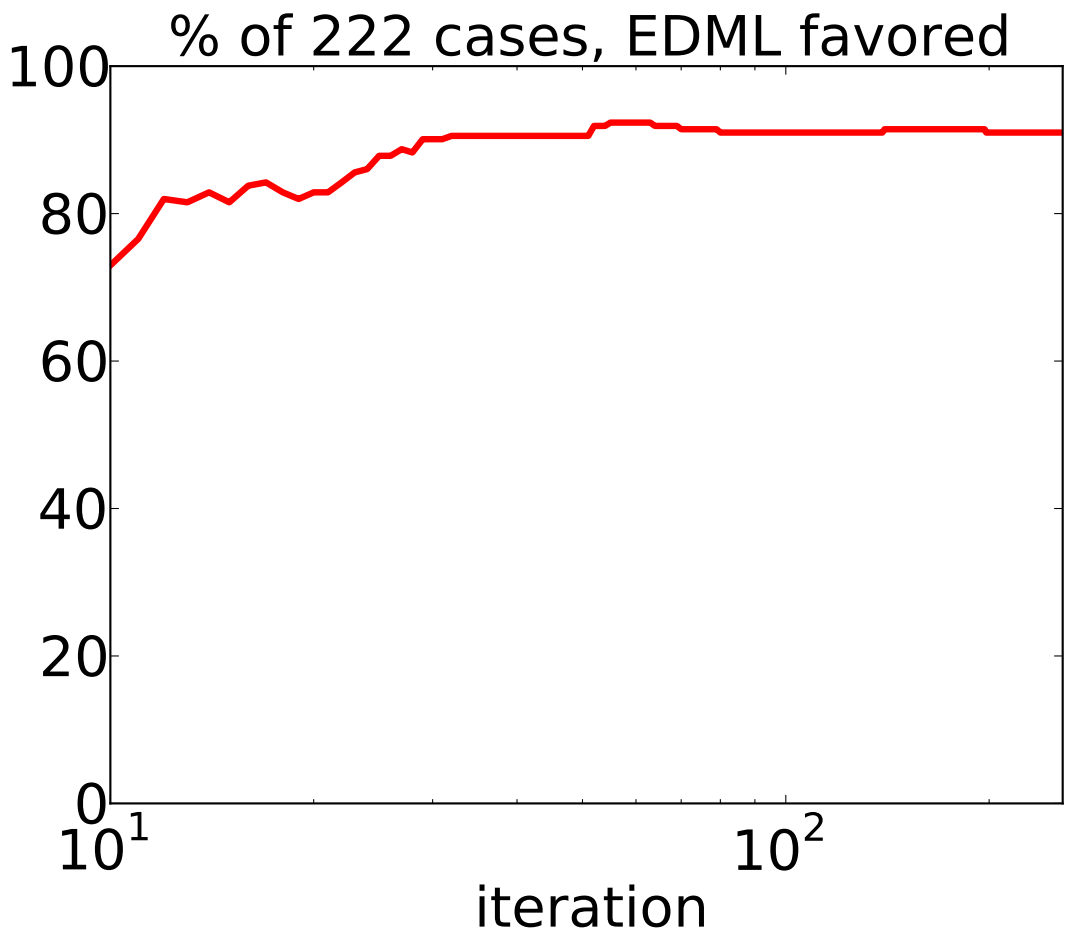

Figure 8: Quality of EDML estimates over 74 networks (3 cases each) induced from binary haplotype data. Going right on the $x$-axis, we have increasing iterations. Going up the $y$-axis, we have an increasing percentage of instances where EDML's estimates were no worse than those given by EM.

would be preferred in practice as well.

# 8 FUTURE AND RELATED WORK

EM has played a critical role in learning probabilistic graphical models and Bayesian networks (Dempster et al., 1977; Lauritzen, 1995; Heckerman, 1998). However learning (and Bayesian learning in particular) remains challenging in a variety of situations, particularly when there are hidden (latent) variables; see, e.g., (Elidan, Ninio, Friedman, & Shuurmans, 2002; Elidan & Friedman, 2005). Slow convergence of EM has also been recognized, particularly in the presence of hidden variables. A variety of techniques, some incorporating more traditional approaches to optimization, have been proposed in the literature; see, e.g., (Thiesson, Meek, & Heckerman, 2001).

Variational approaches are an increasingly popular formalism for learning tasks as well, and for topic models in particular, where variational alternatives to EM are used to maximize a lower bound on the log likelihood (Blei, Ng, & Jordan, 2003). Expectation Propagation also provides variations of EM (Minka & Lafferty, 2002) and is closely related to (loopy) belief propagation (Minka, 2001).

Our empirical results have been restricted to a preliminary investigation of the convergence of EDML, in contrast to EM. A more comprehensive evaluation is called for in relation to both EM and other approaches based on Bayesian inference. We have also focused this paper on binary variables. EDML, however, generalizes to multivalued variables since edge deletion does not require a restriction to binary variables and the key result of Section 4 also generalizes to multivalued variables. The resulting formulation is less transparent though when compared to the binary case since Bayes factors no longer apply directly and one must appeal to a more complex method for quantifying soft evidence; see (Chan & Darwiche, 2005). We expect our future work to focus on a more comprehensive empirical evaluation of EDML, in the context of an implementation that uses multivalued variables. Moreover, we seek to identify additional properties of EDML that go beyond convergence.